\documentclass{article}

    \PassOptionsToPackage{numbers, sort&compress}{natbib}

     \usepackage[preprint]{neurips_2023}

\usepackage[utf8]{inputenc} 
\usepackage[T1]{fontenc}    
\usepackage{hyperref}       
\usepackage{url}            
\usepackage{booktabs}       
\usepackage{amsfonts}       
\usepackage{nicefrac}       
\usepackage{microtype}      
\usepackage{xcolor}         

\usepackage{color,soul}
\usepackage{algorithm}
\usepackage{algorithmic}
\usepackage{amsmath}
\usepackage{amsfonts}
\usepackage[capitalize,noabbrev]{cleveref}
\usepackage{amssymb}
\usepackage{mathtools}
\usepackage{amsthm}
\usepackage{standalone}
\usepackage{graphicx}
\usepackage{tikz}
\usepackage{xurl}
\usetikzlibrary{positioning}
\usetikzlibrary{decorations.pathreplacing}
\newcommand{\indep}{\perp \!\!\! \perp}

\title{Adversarial Distribution Balancing for Counterfactual Reasoning}

\author{%
  Stefan Schrod \\
  Department of Medical Bioinformatics\\
  University Medical Center G\"{o}ttingen\\
  G\"{o}ttingen, Germany\\
  \texttt{stefan.schrod@bioinf.med.uni-goettingen.de} \\
  \And
  Fabian Sinz \\
  Institute for Computer Science and Campus Institute for Data Science (CIDAS)\\ 
  University of G\"{o}ttingen\\
  G\"{o}ttingen, Germany\\
  \texttt{sinz@uni-goettingen.de} \\
  \And
  Michael Altenbuchinger \\
  Department of Medical Bioinformatics\\
  University Medical Center G\"{o}ttingen\\
  G\"{o}ttingen, Germany\\
  \texttt{michael.altenbuchinger@bioinf.med.uni-goettingen.de} \\
}

\begin{document}

\maketitle

\begin{abstract}
	The development of causal prediction models is challenged by the fact that the outcome is only observable for the applied (factual) intervention and not for its alternatives (the so-called counterfactuals); in medicine we only know patients' survival for the administered drug and not for other therapeutic options. Machine learning approaches for counterfactual reasoning have to deal with both unobserved outcomes and distributional differences due to non-random treatment administration. Unsupervised domain adaptation (UDA) addresses similar issues; one has to deal with unobserved outcomes -- the labels of the target domain -- and distributional differences between source and target domain. We propose Adversarial Distribution Balancing for Counterfactual Reasoning (ADBCR), which directly uses potential outcome estimates of the counterfactuals to remove spurious causal relations. We show that ADBCR outcompetes state-of-the-art methods on three benchmark datasets, and demonstrate that ADBCR's performance can be further improved if unlabeled validation data are included in the training procedure to better adapt the model to the validation domain. 
\end{abstract}

\section{Introduction}
	Establishing rules for decision making is a challenging problem in various domains such as precision medicine, social science and marketing. In medicine, for instance, selecting the best intervention is a trade off between the expected benefit and potential side effects, and needs to take into account patient's characteristics such as age, life style, and molecular variables. The development of models to guide this decision process is challenged by the fact that only the outcome for the selected intervention (the factual outcome) is observable while the outcome for unexplored interventions, also referred to as counterfactuals, remains hidden. In the potential outcome framework \cite{Rubin2005}, the conditional average treatment effect (CATE) is defined as the difference between observable and unobservable outcomes and thus provides a rationale for decision making. However, since the latter outcomes are not available in the training data, they need to be inferred on the basis of potential outcome estimates. 
		
	Reliable CATE estimates require an understanding of the underlying data generating process. Non-randomized treatment administration, as common for observational trials, and confounding effects of variables that are causal for both the treatment decision and the outcome, can bias CATE estimates \cite{Pearl2009}. Thus, controlling spurious causal effects and the induced systematic differences in the various treatment populations is a major challenge for CATE estimators. Throughout this article, we make the strong ignorability assumption to guarantee that the differences in CATE are solely due to the causal effect of the intervention itself given the context of the patient. In other words, we assume that the data are unconfounded and that the generating distributions of the various treatment groups sufficiently overlap in order to make the treatment effect identifiable \cite{Pearl2009}. Then, the CATE can be estimated as the difference in potential outcomes \cite{Rubin2005}. Further, assuming strong ignorability enables a balancing of the factual and counterfactual generating distributions to mitigate treatment selection biases, as suggested for instance for deep learning based approaches \cite{Shalit2017}.
	
	Closely related to common balancing procedures of counterfactual reasoning is Unsupervised Domain Adaptation (UDA), which attempts to match the generating distributions of different domains \cite{Ganin2015}. Typically, a predictive model is trained on labelled source data and is subsequently adapted to an unlabelled target domain. Rather than directly reweighting the generating distributions with fixed metrics, adversarial training procedures have been shown to be advantageous for UDA \cite{Ganin2015, Tzeng2017, Saito2018}. 
	In particular, \citet{Ozery-Flato2018} and \citet{bica2020estimating} proposed a Domain-Adversarial Neural Network (DANN) \cite{ganin2016domain} based adversary for CATE predictions, where an independent domain discriminator was used to reweight the factual distributions. Importantly, these approaches aim to completely match distributions between treatments. From UDA, it is understood that this potentially generates ambiguous features near decision boundaries \cite{Saito2018} and, thus, similar issues can be expected for CATE estimation. In particular, treatment biased information can also be highly predictive and as such over-regularizing distributional similarity might compromise the CATE estimates.
	
	We identify factual and counterfactual data with the source and target domain, respectively, and propose Adversarial Distribution Reweighting for Counterfactual Reasoning (ADBCR). ADBCR extends the adversarial UDA training procedure by \citet{Saito2018} to learn continuous outcomes for multiple treatments in a representation learning framework. Accordingly, we define a discriminative distance between two adjacent data representations which is maximized to detect counterfactual samples far away from the support of factual data and subsequently minimize it to reweight a shared data generator. Therefore, ADBCR's objective considers the differences in counterfactual outcome estimates rather than general distributional differences, meaning that the discriminative distance is used to evaluate the support of the counterfactuals. Model selection is based on a criterion which combines the factual objective with the discriminative distance. We further show that the discriminative distance approximates and optimizes an Integral Probability Metric (IPM) representation in each training step, where the space of possible IPM functions is confined to all Deep Neural Network (DNN) representations of the discriminative distance. The UDA motivated adversary can also facilitate training with both factual and completely unlabelled data to further improve the out-of-sample predictions; we refer to this as Unlabelled Adversarial Distribution Reweighting for Counterfactual Reasoning (UADBCR). Model development requires only gradient calculations of simple residual losses, making (U)ADBCR computationally highly efficient even for large sample and feature spaces.
	
	We show that ADBCR outperforms state-of-the-art CATE predictors on two established semi-synthetic datasets for continuous outcome estimates. We further show that the adaptation of unlabelled data further improves the CATE estimates, especially with respect to the Average Treatment Effect (ATE). We finally investigate in ablation experiments the major design choices regarding the discriminative distance and the proposed model selection criterion.
	
\section{Methodology}
	\subsection{The counterfactual problem}
	Let $\mathcal{X}$ denote the $d$-dimensional space of covariates with $X\in \mathcal{X}$ describing the pre-treatment context of individual patients. Throughout this work we consider a binary treatment choice $T \in \{1,0\}$ with a control group $T=0$ and a treated group $T=1$. Hence, the full space of potential outcomes is given by $\mathcal{Y}_{T=0}\times\mathcal{Y}_{T=1}$, where, however, only a single factual outcome $Y_{X|T}$ is observable. The counterfactual outcomes $Y_{X|1-T}$ remain hidden and have to be inferred based on factual data to reveal the 
	CATE for an individual $x\in X$,
	\begin{align}
	\tau(x):&=\mathbb{E}_{Y|X}\left[Y_{T=1}-Y_{T=0}|X=x\right]\,.
	\end{align}
	Evaluating the CATE provides a rationale for decision making. Note, CATE estimates might differ substantially between individuals as a consequence of $X$. The Average Treatment Effect, in contrast, is defined as $\text{ATE} = \mathbb{E}_{x\sim X}(\tau(x))$ and ignores the patient context, which can compromise individual recommendations.
	
	The main challenge addressed in this work is estimating the CATE in scenarios with non-randomized treatment administration as observed in observational trials. In the latter, treatments are administered based on patients' context leading to systematic differences between treatment and control population $p(X|T)\neq p(X|1-T)$. 
	Correct causal conclusions can only be made if the underlying causal graph is identifiable \cite{Pearl2009}, i.e., data should be unconfounded and generated by the same distribution $p(X)$. Following recent work, we make the strong ignorability assumption \cite{Shalit2017,Johansson2020,Athey2016,Wager2018,Yoon2018,Yao2018,Schrod2022}, consisting of ignorability and overlap. First, ignorability requires the potential outcomes and treatment assignments to be conditionally independent given the context of the patients,
	\begin{align}
		Y_{T=0}, Y_{T=1} \indep T|X\,.
	\end{align}
	Second, overlap assumes a non-zero probability for each treatment to be administered,
	\begin{align}
		\forall x\in X ,\, t\in T\,:\,0<p(t|x)<1\,.
	\end{align}
	Whether strong ignorability holds is not testable in observational settings and has to be justified on the basis of domain knowledge about the underlying data generating process \cite{Pearl2009}. Note, in the presence of unmeasured confounders, manual adjustment without assuming strong ignorability can also strongly bias CATE estimates \cite{Ding2017}.
	
	\subsection{CATE estimation and UDA}
    Both, CATE estimation and Unsupervised Domain Adaptation (UDA) attempt to solve a missing data problem; the former transfers knowledge from the factual to the counterfactual domain while the latter uses labelled source data to make predictions for an unlabelled target domain \cite{Ganin2015}.
    In UDA, general covariate shifts can compromise predictions of models which are solely trained on source data without considering the unlabelled target domain \cite{Ganin2015, Tzeng2017, Saito2018}. To transfer information between both, it is assumed that the covariate shifts do not affect the outcome data, $Y\perp D|X$, with domain label $D\in\{\text{Source, Target}\}$, which is equivalent to the ignorability assumption of causal reasoning.
    Hence, we identify the source domain with the factual ($X|T$) and the target domain with the counterfactual ($X|1-T$) samples, respectively. \citet{Johansson2016} showed that both are equivalent under ignorability. As a consequence, the domains correspond to treatment options and the covariate shifts are attributed to biases in the treatment administration. The latter is further subjected to the overlap assumption.
	
	\subsection{Performance Measures}
    While the performance of UDA can be readily assessed using labeled target data for testing, this is not possible for CATE estimates due to the fundamental problem of causal inference \cite{Holland1986}. Here, we employ as performance measures the Precision in Estimation of Heterogeneous Effect (PEHE) \cite{Hill2011},
	\begin{align}
		\epsilon_{\text{PEHE}}=\mathbb{E}_{x\sim X}\bigl([\tau(x)]-[\hat{\tau}(x)]\bigl)^2\;,
	\end{align}
	with the estimated treatment effect $\hat{\tau(X)}$, and the Average Treatment Effect (ATE),
	\begin{align}
	    \epsilon_{\text{ATE}}=\bigl|\mathbb{E}_{x\sim X}[\tau(x)]-\mathbb{E}_{x\sim X}[\hat{\tau}(x)]\bigl|\;.
	\end{align}
	Both, however, can only be calculated when the counterfactual outcome data is available.

\section{Related Work}
	The practical importance of guiding decision processes based on observational data has lead to the development of a wide variety of CATE estimators,
	comprising single (S-learner) and treatment-specific (T-learner) models \cite{Kuenzel2019}, Inverse Propensity Weighting (IPW) \cite{Rosenbaum1983}, tree \cite{Chipman2010, Athey2016} and forest based estimators \cite{Wager2018}, and deep CATE predictors.
	The latter, which are also the focus of this work, are often based on a representation learning framework \cite{Bengio2013}, in which separate treatment-specific representations are learned based on a shared feature extractor \cite{Shalit2017,Louizos2017,Schwab2018,Yao2018,Hassanpour2019, Hassanpour2019a,Kuenzel2019,Shi2019,Assaad2021,Curth2021a,Schrod2022}. This shared representation is used to reduce effects from biased treatments via balancing the factual $X|T$ and counterfactual $X|1-T$ distributions. Distributional differences can be regularized using Integral Probability Metrics (IPMs) \cite{Shalit2017,Hassanpour2019a,Assaad2021,Schrod2022} defined by 
	\begin{align}
		\text{IPM}_\mathcal{F}&(X|T,X|1-T)=\nonumber \\ \sup_{f\in\mathcal{F}}&\Bigl|\mathbb{E}_{X|T}[f(x)]-\mathbb{E}_{X|1-T}[f(x)]\Bigl|\;, \label{eq:IPM}
	\end{align}
	with a real-valued, bounded and measurable function $f\in\mathcal{F}$ \cite{Mueller1997}.
	In practice, the calculation of the IPM requires a representation such as Maximum Mean Discrepancy \cite{Gretton2012} or Wasserstein Distance \cite{Cuturi2014}.
	Alternative measures were proposed to mitigate treatment biases, including middle-point distance minimization using a position-dependent deep metric (SITE), as introduced by \citet{Yao2018}, Nearest Neighbour (NN) estimates \cite{Schwab2018}, or simultaneously predicting propensity scores based on an additional representation head to balance the treatment distributions \cite{Shi2019}. Adversarial strategies typically rely on an additional domain critic to eliminate treatment bias, i.e., use a DANN \cite{ganin2016domain} adversary for distributional alignment. Specifically, \citet{Ozery-Flato2018} proposed adversarial balancing as a pre-training procedure, \citet{bica2020estimating} analyzed time-dependent treatment effects and used a domain critic at each discrete point in time, and \cite{Kallus2020} theoretically connected the introduced adversarial distance with IPM-based approaches.
	Additionally, \citet{Yoon2018} and \citet{du2021adversarial} proposed Generative Adversarial Networks (GANs) which use artificially generated data to distinguish the domains. These adversarial procedures, however, do not directly consider factual information. As such the alignment is agnostic with respect to the factual outcomes, potentially leading to an uncontrolled loss of factual accuracy due to over-enforced distributional similarity. We pursue an alternative strategy combining factual and counterfactual information in the adversary.

\section{Adversarial Distribution Balancing for Counterfactual Reasoning}
\begin{figure}
		\begin{center}
			\resizebox{0.9\linewidth}{!}{
				\input{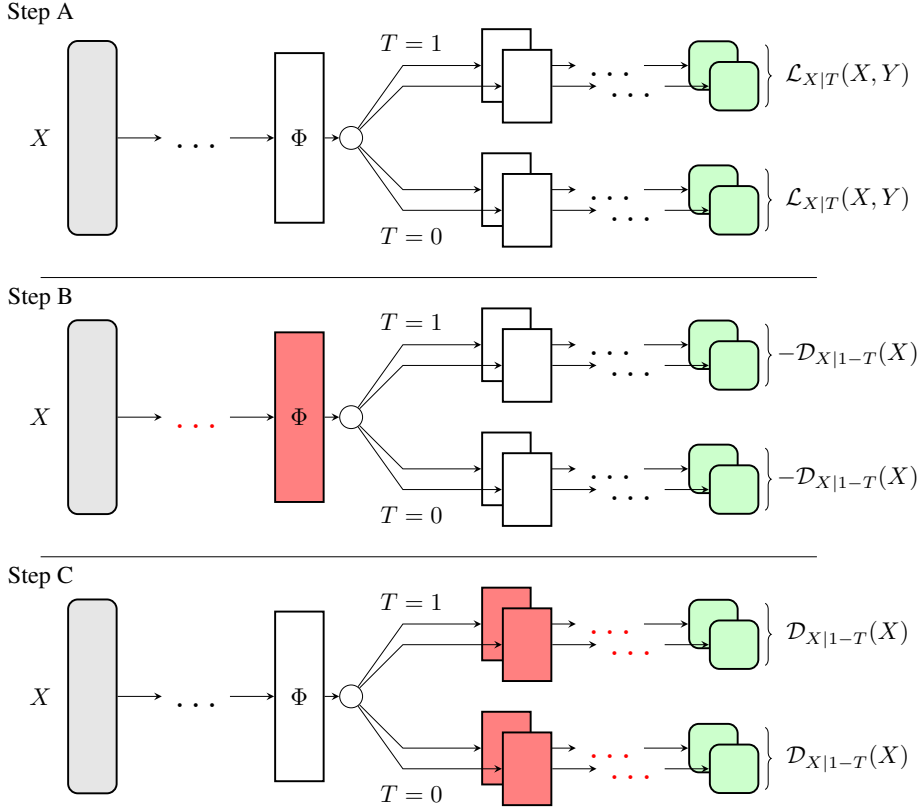}
			}
			\caption{The ADBCR network architecture and the main training steps; \textbf{Step A}: Factual outcome fit, \textbf{Step B}: Maximize the counterfactual discrepancy for the treatment dependent representations \textbf{Step C}: Minimize the counterfactual discrepancy based on shared representation. Note that the parameters of the red layers are fixed for the respective optimization step. \label{fig:CFMCD}}
		\end{center}
	\end{figure}

	We propose Adversarial Distribution Balancing for Counterfactual Reasoning (ADBCR) as a general framework for treatment balanced CATE estimation which is motivated by recent developments in UDA. Typically, UDA attempts to completely match feature distributions between different domains. This can be sub-optimal since this can generate ambiguous features near decision boundaries, and since it can remove domain characteristics relevant for the prediction task \citet{Saito2018}. In the context of causal inference, treatment biased features are often highly predictive for the outcome and, hence, over enforcing similarity is often accompanied by a loss of predictive accuracy.
	To mitigate these issues, \citet{Saito2018} proposed to maximize the discrepancy between two classifiers’ outputs to detect target samples far from the support of the source. ADBCR follows a similar strategy for counterfactual reasoning: we train two distinct, differently initialized treatment-heads $R^0_t(\Phi(X)),R^1_t(\Phi(X)) \in \mathcal{R}$ for each treatment, four in total (see \cref{fig:CFMCD}). All heads are based on the same shared latent representation $\Phi(X)$. ADBCR then aims to transfer knowledge from the factual to the counterfactual treatment by removing the distributional differences induced by non-randomized treatment administrations. To assess if the counterfactuals are properly supported by the factual data distribution, it compares the two distinct treatment heads on the respective counterfactual input data.
	The rationale is that if two distinct outcome heads agree on the factual outcome, then they should also agree on the counterfactual outcome unless they are influenced by spurious dependencies.
	Thus, ADBCR evaluates the agreement of the two treatment dependent outcome heads on the respective counterfactual distributions using the discriminative distance
	\begin{align}
		\mathcal{D}&_{X|1-T}(X):= \sum_t \mathbb{E}_{X|T=1-t}d\bigl( R^0_t(\Phi(X)), R^1_t(\Phi(X))\bigl)\;,
	\end{align}
	with a point-wise metric $\text{d}:\mathbb{R}^n \times \mathbb{R}^n\rightarrow \mathbb{R}$. Throughout this work we use the $l_1$ distance as motivated by \citet{Saito2018}. Note that for a small discrepancy the counterfactual sample might be well supported by the respective factuals and possible treatment administration biases might be mitigated. 
	If the heads disagree on their estimates, the prediction might be compromised.
	To detect distributional dissimilarities and to guarantee distinct representations, we define an adversary that explicitly maximizes the discrepancy based on counterfactual samples with respect to the outcome representations $\mathcal{R}$. Subsequently, matching the support of factual and counterfactual data is enforced by minimizing the discrepancy with respect to the latent representation $\Phi$.
	Altogether, this procedure comprises the three training steps outlined below (see also \cref{alg:alg}) and illustrated in \cref{fig:CFMCD}.
	
	\paragraph{Step A: Factual Fit} In a first step, the model is trained to correctly reproduce the factual outcomes, where the latent representation $\Phi$ and the respective treatment outcome heads $R^r_t\in\mathcal{R}$ are fitted simultaneously. We use the standard Mean Squared Error (MSE) objective 
	\begin{align}
		&\min_{\Phi,\mathcal{R}}\mathcal{L}_{X|T}(X,Y)\\
		\mathcal{L}_{X|T}(X,Y)=&\sum_{t,r} \mathbb{E}_{X|T=t} \left[ \bigl(R^r_t(\Phi(X))-Y\bigl)^2\right]
	\end{align}
	given the factual outcome $Y$ and corresponding context $X$.

	\paragraph{Step B: Counterfactual maximization} Second, we maximize the adversarial discriminative distance between the adjacent representations $R^0_t$ and $R^1_t$ for each treatment $t$, 
	\begin{align}
		&\min_{\mathcal{R}}\; \mathcal{L}_{X|T}(X,Y) - \mathcal{D}_{X|1-T}(X)
	\end{align}
	where the discriminative distance between the representations $\mathcal{D}_{X|1-T}(X)$ is evaluated using data from the respective counterfactual domain $X|1-T$.
	In this step, all parameters which map to the shared representation $\Phi$ are kept fixed (red colored parts of Step B in the architecture \cref{fig:CFMCD}). We additionally weight the adversary against the factual objective as motivated by \cite{Saito2018} in order to prevent an uncontrolled loss of prediction accuracy for the factual outcome. Intuitively, this step induces a difference between data representations $R^0_t$ and $R^1_t$, while controlling the agreement of the factual predictions.
    
	\paragraph{Step C: Counterfactual minimization} This step counteracts the adversary trained in Step B by minimizing the discriminate distance $\mathcal{D}_{X|1-T}(X)$ with respect to the shared representation $\Phi$, while the parameters of the individual outcome heads are kept fixed (red layers in Step C of \cref{fig:CFMCD}).
	\begin{align}
		\min_{\Phi} \mathcal{D}_{X|1-T}(X)
	\end{align}
	This step is repeated $k$ times to enforce better agreement between the representations $R_t^{0}$ and $R_t^{1}$.
	
	The rational behind step B and C is that if the two individual representations $R_t^{0}$ and $R_t^{1}$ agree in their factual and counterfactual prediction based on the learned data representation $\Phi$, then the model is likely to generalize and transfer meaningful knowledge to the counterfactual domain. The final predictions for the potential outcomes are averages of the adjacent outcome heads for each treatment, respectively.
	The full training procedure is given in \cref{alg:alg}. We use a fully connected feed forward architecture with dropout layers and update the parameters for each step individually using the Adam optimizer \cite{Kingma2014}. Our implementation of ADBCR is publicly available at \url{https://github.com/sschrod/ADBCR}. 
	
	\begin{algorithm}[t]
		\caption{ADBCR}
		\label{alg:alg}
		\textbf{Input}: Factual samples $(x_i,t_i,y_i)\in X\times T \times Y$, shared network $\Phi$ with parameters $\theta$, treatment dependent representation networks $R_t^r$ with parameters $\rho_t^r$, number of adversarial steps $k$.
		\begin{algorithmic}[1]
		\WHILE{Early stopping for $X\in X_{\text{val}}$ is not met:\\ \quad$\mathcal{L}_\text{val}=\mathcal{L}_{X|T}(x_i,y_i) + \mathcal{D}_{X|1-T}(x_i)$}
		\FOR{$X$ \textbf{in} $X_\text{train}$}
	    \STATE \textbf{STEP A:} Update parameters of $\Phi$ and $R_t^r$ by\\ \quad$\nabla_{\theta,\rho_t^r}\mathcal{L}_{X|T}(x_i,y_i)$\\
		\STATE \textbf{ STEP B:} Update parameters of $R_t^r$ by\\
			\quad$\nabla_{\rho_t^r}\mathcal{L}_{X|T}(x_i,y_i) - \mathcal{D}_{X|1-T}(x_i)$\\
		\STATE \textbf{ STEP C:} Update parameters of $\Phi$ by\\
		\FOR{$i=0$ \textbf{to} k}
		\STATE $\nabla_{\theta} \mathcal{D}_{X|1-T}(x_i)$
		\ENDFOR
		\STATE \textbf{ STEP A:} Update parameters of $\Phi$ and $R_t^r$ by\\ \quad$\nabla_{\theta,\rho_t^r}\mathcal{L}_{X|T}(x_i,y_i)$
		\ENDFOR
		\ENDWHILE
		\end{algorithmic}
	\end{algorithm}
	
	\paragraph{UADBCR}
	The adversarial distribution balancing procedure of ADBCR is not limited to the adaptation of counterfactual samples, but can also adapt to unlabeled data without observed outcome. To achieve this, the counterfactual adversary (steps B and C) additionally evaluate the unlabelled data for both treatment domains, simultaneously. In practice this could be used to include an inductive bias towards the test cohort, e.g., to facilitate predictions for patients with different treatment histories by simultaneously eliminating distributional differences to this population. For ADBCR that means, the unlabelled data are considered as counterfactual for each treatment choice and batches are sampled from a mixture of factual and unlabelled data, which is facilitated by the treatment dependent adversaries. We refer to this as Unlabelled Adversarial Distribution Balancing for Counterfactual Reasoning (UADBCR).

	\paragraph{The discriminative distance}
	The calculation of the distance $\mathcal{D}_{X|1-T}(X)$ requires an appropriate point-wise metric $d$. In \citet{Saito2018}, the $l_1$ residual distance was used inspired by both empirical results and the theoretical arguments given by \citet{Ben-David2010}. Here, we follow the same strategy and additionally compare it to the MSE disciminative distance in ablation experiments. Moreover, we illustrate a relationship to IPMs in the following. IPMs (\cref{eq:IPM}) evaluate the distributional differences of the domains given an optimal critic function $f\in \mathcal{F}$ and provide an upper bound for the expected error of CATE estimates \cite{Shalit2017}. For empirical calculations, $\mathcal{F}$ has to be sufficiently general in order to uniquely identify whether the domains are equal \cite{Gretton2012}.
	Popular IPM choices such as the MMD or the Wasserstein Distance require a predefined kernel or evaluate pair-wise distributional differences using a fixed distance measure, respectively. Both might not be sufficient to capture all distributional differences \cite{Kallus2020}. Therefore, ADBCR dynamically approximates $f$ during the optimization procedure, where the space of functions $\mathcal{F}$ is confined by the network structure and the discrepancy measure, i.e., $\mathcal{F}: R_t^0\times R_t^1\rightarrow \mathbb{R}$ is the space of all discriminative distances given by the two adjacent DNN representations $R_t^0$ and $R_t^1$. The corresponding IPM between factual $(X|T)$ and counterfactual $(X|1-T)$ domain is given by
	\begin{align}
		\text{IPM}&_\mathcal{F}(X|T,X|1-T)=\sup_{\mathcal{D}\in\mathcal{F}}\Bigl|\mathcal{D}_{X|T}(X)-\mathcal{D}_{X|1-T}(X)\Bigl|\label{eq:ipm}\,,
	\end{align}
	with $\mathcal{D}_{X|T}(X)$ defined analogously to $\mathcal{D}_{X|1-T}(X)$ but conditioned on the factual treatment $T$.
	Since $\mathcal{D}_{X|T}(X)$ is inherently minimized in Step A, where $R_t^0$ and $R_t^1$ are both trained to reproduce the factual outcomes, the term $\mathcal{D}_{X|T}(X)$ is assumed to be very small. Therefore, we approximate \cref{eq:ipm} by
	\begin{align}
	    \max_{\mathcal{D}\in\mathcal{F}}\Bigl|\mathcal{D}_{X|1-T}(X)\Bigl| \label{eq:ADBCR_max}\,,
	\end{align}
	where the adversary relies on counterfactual data, only.
	In summary, we solve a mini-max problem, where $\mathcal{D}_{X|1-T}$ in \cref{eq:ADBCR_max} is maximized in Step B and subsequently minimized in Step C. Importantly, this uses predictive information to build the IPM representation and is computationally much cheaper to calculate than fixed IPM measures such as the Wasserstein Distance or the MMD since (U)ADBCR only requires gradients for $l_1$ and MSE losses.

    \paragraph{Model selection}
	The accuracy of CATE estimates strongly depends on an appropriate balance between factual accuracy and distributional similarity \cite{Hassanpour2019a, Zhang2020, Assaad2021}. This trade-off is challenged by missing counterfactual information, which prevents standard model selection procedures such as cross-validation. On the other hand, evaluating the factual error \cite{Yao2018, Schrod2022} does not account for the complex structure of the underlying data generating process and might bias CATE estimates towards factual accuracy.
	A popular remedy is to use imputed CATE values of an independent Nearest Neighbour (NN) estimator, also known as $\text{PEHE}_{\text{NN}}$ \cite{Shalit2017,Schwab2018,Johansson2020,Hassanpour2019}. \citet{Assaad2021} compared different imputation methods and in general model selection strongly depends on the quality of the imputation estimator, which is likely to be compromised in high dimensional settings for NN \cite{Indyk1998}.
	
	We propose to select hyper-parameters based the intrinsic objective of ADBCR, i.e., the combination of Step A and C,
	\begin{align}
	    \mathcal{L}_{\text{val}} = \mathcal{L}_{X|T}(X,Y) + \mathcal{D}_{X|1-T}(X)\label{eq:val}\,,
	\end{align}
	This enforces factual accuracy and supports the counterfactual samples. Note, different choices of discriminative measure or scaling the two losses of \cref{eq:val} can further dis- or encouraged balance between the generating distributions. We investigate the former in ablation experiments.

	\begin{table*}[t]
        \centering
        \caption{Within-sample and out-of-sample results of the error of PEHE and ATE estimates for IHDP and NEWS data.}
        \vspace{5pt}
        \label{tab:results}
        \resizebox{\linewidth}{!}{%
			\begin{tabular}{lrrrrrrrr}
				\toprule
				&\multicolumn{4}{c}{IHDP} 						  & \multicolumn{4}{c}{NEWS}\\
				\cmidrule(r){2-5}\cmidrule(l){6-9}
				&\multicolumn{2}{c}{Within-Sample} 						  & \multicolumn{2}{c}{Out-of-sample}&\multicolumn{2}{c}{Within-Sample} 						  & \multicolumn{2}{c}{Out-of-sample}\\
				\cmidrule(r){2-3}\cmidrule(lr){4-5}\cmidrule(lr){6-7}\cmidrule(l){8-9}
				&$\sqrt{\epsilon_{\text{PEHE}}}$ & $\epsilon_{\text{ATE}}$ & $\sqrt{\epsilon_{\text{PEHE}}}$ & $\epsilon_{\text{ATE}}$ &$\sqrt{\epsilon_{\text{PEHE}}}$ & $\epsilon_{\text{ATE}}$ & $\sqrt{\epsilon_{\text{PEHE}}}$ & $\epsilon_{\text{ATE}}$\\
				\midrule
				S-Lasso & $5.64\pm .83$ & $.83\pm .16$ & $5.58\pm .81$ & $1.1\pm .21$& $4.23\pm .17$ & $2.5\pm .07$ & $4.25\pm .17$ & $2.5\pm .07$\\
				T-Lasso & $2.09\pm .34$ & $\mathbf{.12\pm .01}$ & $2.02\pm .31$ & $.25\pm .04$& $2.03\pm .08$ & $.33\pm .02$ & $2.31\pm .16$ & $.34\pm .03$\\
				TARNet & $0.93\pm .07$ & $.25\pm .03$ & $0.91\pm .04$ & $.26\pm .03$& $1.81\pm.05$ & $.32\pm.04$ & $1.93\pm.06$ & $.30\pm.04$\\
				CFRNet & $0.83\pm .07$ & $.26\pm .03$ & $0.81\pm .05$ & $.27\pm .04$& $1.83\pm.05$ & $.34\pm.04$ & $1.98\pm.06$ & $.37\pm.04$\\
				SITE & $0.93\pm .06$ & $.30\pm .04$ & $0.90\pm .06$ & $.30\pm .04$& $2.20\pm.07$ & $.18\pm.02$ & $2.44\pm.09$ & $.22\pm.03$\\
				ABCEI & $0.79\pm .07$ & $\mathbf{.12\pm .02}$ & $0.83\pm .07$ & $\mathbf{.15\pm .03}$& $1.63\pm.05$ & $.18\pm.03$ & $1.81\pm.07$ & $.23\pm.04$\\
				DANNCR & $0.93\pm .10$ & $.26\pm .05$ & $1.06\pm .13$ & $.26\pm .05$& $1.42\pm.04$ & $\mathbf{.10\pm.02}$ & $1.64\pm.05$ & $\mathbf{.12\pm.01}$\\
				\midrule
				\bfseries ADBCR & $0.70\pm .07$ & $.17\pm .02$ & $0.75\pm .07$ & $.20\pm .03$& $1.40\pm .06$ & $.22\pm .02$ & $1.59\pm .06$ & $.25\pm .02$\\
				\bfseries UADBCR & $\mathbf{0.69\pm .06}$ & $.15\pm .01$ & $\mathbf{0.73\pm .07}$ & $.17\pm .02$&  $\mathbf{1.37\pm .06}$ & $.22\pm .03$ & $\mathbf{1.58\pm .06}$ & $.22\pm .02$\\
				\bottomrule
			\end{tabular}
		}
    \end{table*}

	\section{Experiments}
	The performance assessment of CATE estimates for both data associated with a single factual outcome (in-sample), and data not associated with an observable outcome (out-of-sample), is a non-trivial task since we never observe the CATE directly. Consequently, continuous benchmark data are based on real covariates and treatments, while the outcomes and treatment imbalances are simulated using a predefined data generating process. Such artificially created data do usually not represent the full complexity of real world data \cite{Curth2021}. To ensure a fair comparison we do not impose any specifically designed priors which could bias results towards the selected data generating process. We investigate two linear baselines; a single learner Lasso (S-Lasso) and a treatment dependent Lasso (T-Lasso), and compare ADBCR and UADBCR to the state-of-the-art deep models, TAR- and CFRNet regularized with Wasserstein Distance \cite{Shalit2017,Johansson2020}, SITE \cite{Yao2018}, ABCEI \cite{du2021adversarial}. Moreover, we implemented a DANN Counterfactual Reasoning model (DANNCR) inspired by the ideas of \citet{Ozery-Flato2018} and \citet{bica2020estimating}, since these methods are closely related to ADBCR although developed for other applications, i.e., for general preprocessing of treatment biased data and for time-dependent treatment effects, respectively. DANNCR uses a domain discriminator with a cross-entropy objective to predict whether the samples were generated by the treated or control population in order to remove distributional differences. We implemented the same network architecture as for ADBCR but with a single outcome head per treatment, and screened the same space of hyper-parameters. The best model for DANNCR is selected based on factual accuracy for the validation data as suggested in the original work \cite{bica2020estimating}. In addition to the main baselines we further compared (U)ADBCR to Meta-Learning models on an adapted version of the Infant Health Development Program (IHDP) \cite{Curth2021a} in the Supplementary Materials. The comparison includes the DragonNet model by \citet{Shi2019}, regression adjustment, propensity weighting and doubly robust models as evaluated by \citet{Curth2021a}. For additional implementation details refer to the Supplementary Materials.

	\subsection{Datasets}
	ADBCR is designed to make continuous CATE predictions and we therefore compare three continuous outcome baseline datasets:
	\paragraph{IHDP}
	Likely the most popular semi-synthetic CATE benchmark is based on a randomized trial conducted by the Infant Health Development Program (IHDP), studying the impact of specialist child care to infants with low birth weight based on cognitive scores \cite{Brooks-Gunn1992}. The dataset comprises $25$ clinical parameters describing the pre-treatment context of the mothers and the infants, where $139$ of the $747$ infants received specialist child care $(T=1)$.
	To mimic an observational trial and to introduce selection bias, \citet{Hill2011} removed a specific proportion of the treated individuals (children with non-white mothers) and simulated a non-linear response (Setting "B"). We report the results for the first $100$ outcome realizations for the same $63/27/10$ train/validation/test splits as initially used by \citet{Shalit2017}.
	
	Note, IHDP is commonly reported based on a different number of realizations and therefore, we additionally evaluate (U)ADBCR for $10$ and $1000$ realizations and compare it to the originally reported results of \citet{Shalit2017} and \citet{Yao2018} in the supplementary materials.
	
	\paragraph{NEWS}
	As a second baseline, we use the NEWS dataset introduced by \citet{Johansson2016}. It models the reader opinion of $5000$ randomly sampled news articles from the New York Times based on the viewing device, i.e., desktop $(T=0)$ or mobile $(T=1)$. Each article is described by a word count vector of $3477$ individual words based on the $100$ most common words of each topic. Treatments are distributed according to device preferences for certain topics given $50$ different realizations. We refer to \citet{Johansson2016} for a detailed description of the data generating process.

\begin{table}[t]
    \centering
    \caption{Within-sample and out-of-sample results of the error of PEHE and ATE estimates for the ACIC2016 data.}
		\begin{tabular}{lrrrr}
			\toprule
			&\multicolumn{4}{c}{ACIC2016} \\
			\cmidrule(r){2-5}&\multicolumn{2}{c}{Within-Sample}& \multicolumn{2}{c}{Out-of-sample}\\
			\cmidrule(r){2-3}\cmidrule(lr){4-5}
			&$\sqrt{\epsilon_{\text{PEHE}}}$ & $\epsilon_{\text{ATE}}$ & $\sqrt{\epsilon_{\text{PEHE}}}$ & $\epsilon_{\text{ATE}}$\\
			\midrule
		    S-Lasso & $3.31\pm .66$ & $.34\pm .08$ & $3.30\pm .67$ & $.30\pm .11$\\
		    T-Lasso & $2.51\pm .44$ & $.38\pm .11$ & $2.62\pm .44$ & $.54\pm .13$\\
		    TARNet & $2.07\pm .39$ & $.51 \pm .17$ & $2.05 \pm .39$& $.37\pm .13$\\
		    CFRNet & $2.13\pm .45$ & $.28 \pm .11$& $2.13 \pm .44$ & $.40 \pm .15$\\ 
		    SITE & $2.17 \pm .44$ & $.36\pm .12$ & $2.23\pm .42$ & $ .47 \pm .20$\\
		    ABCEI & $2.19 \pm .47 $ & $\mathbf{.16 \pm .06}$ & $2.34 \pm .48$ & $ .23 \pm .09$\\
		    DANNCR & $2.77 \pm .53$ & $.33\pm .05$ & $2.72\pm .51$ & $.38 \pm .07$ \\
			\midrule
			ADBCR & $\mathbf{1.73\pm .33}$ & $.24 \pm .05$ & $\mathbf{1.81\pm .33}$ & $\mathbf{.18 \pm .06}$ \\
			UADBCR & $1.75\pm .32$ & $.18\pm .03$  &$1.90\pm .32$ & $.21\pm .06$\\
			\bottomrule
		\end{tabular}
	\label{tab:ACIC2016_results}
\end{table}
	
	\paragraph{ACIC2016}
	We further use 10 representations of the 2016 Atlantic Causal Inference Competition (ACIC2016) \cite{dorie2019automated}. The data was particularly designed to cover different data generating processes to not favour specifically designed models. The used representations were obtained from \href{https://github.com/BiomedSciAI/causallib/tree/master/causallib/datasets/data/acic_challenge_2016}{https://github.com/BiomedSciAI/causallib}.
	
	\subsection{Results}
	The in-sample and out-of-sample results obtained on the IHDP, NEWS and ACIC2016 data are presented in \cref{tab:results} and \cref{tab:ACIC2016_results}. For the linear baselines, we observe that decoupling the treatment and the covariates and utilizing treatment dependent learners significantly increases the performance (compare S versus T-Lasso). In particular, the T-Lasso shows very competitive ATE performance across all settings, and even outperforming all competing models for in-sample IHDP ATE estimates. However, the results also show that capturing population level effects, i.e. ATE, does not correlate with well calibrated individual estimates. In particular the T-Lasso on IHDP shows that predicting ATE can be can be highly efficient even though patients can not be properly discriminated. As a consequence the linear baselines are consistently outperformed by the deep models with respect to PEHE scores.
	
    For IHDP data, we observe that Wasserstein balancing (CFRNet) significantly improves the individual predictions of TARNet while showing comparable ATE performance. This is in line with the originally reported results obtained on 1000 realizations of the data, where ADBCR shows competitive performance (for the full results, refer to supplementary materials). SITE and DANNCR, in contrast, were not able to improve the TARNet predictions (for a performance comparison of SITE to the originally reported results obtained on 10 realizations refer to the Supplementary Materials). ABCEI shows the best performing ATE estimates but such as all other baselines it is consistently out-competed by ADBCR with respect to the individual CATE estimates. Including additional, unlabelled data (UADBCR) further improves ADBCR estimates, particularly with with respect to ATE.
    
    For the NEWS data, we observe that neither CFRNet nor SITE are able to improve on the individual estimates of the unbalanced TARNet. The adversarial models on the other hand improve the individual estimates significantly, which we attribute to the increased flexibility of the adversaries to detect and regularize distributional differences on such high-dimensional data. Overall, ADBCR outcompetes all baselines with respect to the individual estimates.

    On the ACIC2016 data, the unbalanced TARNet even outperforms the adversarial baselines ABCEI and DANNCR with respect to PEHE score on both in-sample and out-of-sample predictions. On this data, solely ADBCR provides distributional balancing which improves the predictions with respect to the unbalanced TARNet baseline. On all three datasets ADBCR consistently provides the best individual estimates. These are further improved if unlabeled data are included in the training procedure (UADBCR) for IHDP and NEWS, while ADBCR and UADBCR performed similar on ACIC2016.
    
	\begin{table}[t]
	        \centering
	        \caption{ADBCR ablation study results for within- and out-of-sample estimates based 100 realizations of the IHDP data.}
			\begin{tabular}{lrr}
				\toprule
				& \multicolumn{2}{c}{Out-of-sample}\\
				&$\sqrt{\epsilon_{\text{PEHE}}}$ & $\epsilon_{\text{ATE}}$\\
				\midrule
				A-TARNet & $1.07\pm .14$ & $0.25\pm .04$ \\
				ADBCR $\text{PEHE}_\text{NN}$  & $3.15\pm .47$ & $0.67\pm .11$ \\
				ADBCR $\mathcal{L}_{\text{val},X|T}$  & $0.77\pm .08$ & $\mathbf{0.17\pm .02}$\\
				ADBCR $\mathcal{D}^2_{X|1-T}$ & $0.80\pm .08$ & $0.18\pm .02$ \\
				\bfseries ADBCR  & $\mathbf{0.75\pm .07}$ & $0.20\pm .03$ \\
				\bottomrule
			\end{tabular}
		\label{tab:ablation_study}
	\end{table}

	\section{Ablation Study}
	To further motivate our design choices, we investigated different discriminative distances and validation measures in ablation experiments to study their effect on ADBCR's performance. As a baseline, we removed the adversarial training steps B and C. This approach simply extends the TARNet architecture to an ensemble learner where a second outcome head is added for each treatment; the Adapted TARNet (A-TARNet). We compare this to modified models utilizing the full training procedure. This includes $\text{PEHE}_\text{NN}$ validation as introduced by \citet{Shalit2017}, simple factual evaluation $\mathcal{L}_{\text{val},X|T}$ and a model that utilizes a quadratic discriminative distance
	\begin{align}
	\mathcal{D}^2_{X|1-T}(X):= \sum_t \mathbb{E}_{X|T=1-t}\bigl( R^0_t(\Phi(X)) - R^1_t(\Phi(X))\bigl)^2
	\end{align}
	for training and evaluation.
	
	\cref{tab:ablation_study} reports the out-of-sample results. We observe that adding adversarial balancing to the A-TARNet architecture significantly improves the CATE predictions. Model selection using the $\text{PEHE}_\text{NN}$ estimator showed highly unfavorable performance for ADBCR, as it was also the case for using a quadratic discrepancy measure. In the latter case, the reduced performance could be the consequence of over-enforcing distributional similarity. This effect might be substantially weakened for the linear discrepancy of ADBCR, leading to improved individual estimates.

	\section{Conclusion}
	In this article, we have introduced ADBCR, a novel deep representation learning method for unbiased CATE estimation. ADBCR utilizes an adversarial training procedure to efficiently identify and balance distributional differences to mitigate treatment biases. Experimental results show that ADBCR outperforms both state-of-the-art representation learners and previously proposed adversarial CATE predictors.
	This is even more significant on high-dimensional data, where fixed balancing measures are not able to capture the rich structure of the data \cite{Kallus2020}. Adversarial learning, on the other hand, efficiently scales to large-scale problems while remaining computationally efficient. Our proposed model selection criterion improves model performance and facilitates UDA learning via unlabelled validation samples which can be included into the training procedure. Hence, UADBCR simultaneously balances the treatment bias and adapts validation data without observable outcome.
	Both the architecture and training procedure of (U)ADBCR are highly flexible and can be straightforwardly extended to multiple treatments. Moreover, it could be adapted to domain adaptation with partially labelled domains.
	
	\paragraph{Limitations}
	The evaluation of counterfactual models in real-world applications is usually challenged by the fact that the counterfactual outcome is not available. As a consequence, model selection is typically based on factual accuracy only or proxies like NN PEHE estimates. In contrast, ADBCR's model selection criterion also accounts for the distributional alignment to reduce the bias of CATE estimates towards factual accuracy. Although we could empirically show that this procedure is highly efficient and beneficial for model performance, additional domain knowledge might be necessary to fine-tune between factual accuracy and distributional alignment. Moreover, ADBCR, as well as all other comparable methods, assume strong ignorability. When strong ignorability is violated, removing distributional differences induced by confounding factors might compromise the CATE estimates. 
	
	\section*{Acknowledgments and Funding}
	This work was supported by the German Federal Ministry of Education and Research (BMBF) [01KD2209D] and within the e:Med research and funding concept [01ZX1912C].

	\bibliographystyle{plainnat}
    \bibliography{ADBCR}

    \newpage
    \section*{Supplementary Materials}
	\section*{Implementation Details}
	\paragraph{LASSO} Both the T- and S-learner implementations are based on the cross-validated Lasso implemented in the sklearn package \cite{Pedregosa2011}. This provides a factual fit to the data. To retrieve counterfactual estimates, we vary the treatment parameter or evaluate the respective treatment specific model. We evaluated a range of regularization parameters, $\alpha \in \{10^k\}_{k=-3}^{2}$, for each of the methods.
	
	\paragraph{TAR- and CFRNet} We used the original implementation of \citet{Shalit2017} provided at \url{https://github.com/clinicalml/cfrnet}. For the results presented in the main text, we used their proposed set of hyper-parameters for TAR- and CFRNet, i.e., $3$ input layers of size $200$, 3 output layers of size $100$, ELU activation functions, a batch size of $100$ and $p_\alpha=0.3$. The latter regularizes the strength of the Wasserstein distance balancing for CFRNet. Additionally, we sampled and evaluated $25$ sets of hyper-parameters. Although we obtained smaller validation $\text{PEHE}_{\text{NN}}$ scores, these are associated with poor counterfactual performance ($\sqrt{\epsilon_{\text{PEHE}}}=1.91 \pm 0.30$). This might explain the inferior performance of ADBCR using NN-evaluation and the NN-estimates.
	The parameters used for the NEWS and ACIC2016 model selection are shown in \cref{Stab:CFRparameters}. Note, ACIC2016 uses the same hyper-parameter search as initially proposed by \citet{Shalit2017}. For TAR- and CFRNet, we evaluated $25$ randomly sampled hyper-parameter combinations for the $50$ realizations of the data. Note, $p_\alpha=0$ corresponds to the TARNet model.
	\begin{table}[h]
	    \centering
		\caption{Hyper-parameter ranges for NEWS for TARNet, CFRNet, SITE and ABCEI}
		\label{Stab:CFRparameters}
					\begin{tabular}{ll}
					    \toprule
						Parameter & Range\\
						\midrule
						$p_\alpha$ (only for CFRNet) & $\{10^k\}_{k=-5}^3$\\
						$p_\text{PDDM}$ (only for SITE) & $\{10^{k/2}\}_{k=-2}^6$\\
						$p_\text{MPDM}$ (only for SITE)& $\{10^{k/2}\}_{k=-2}^6$\\
						$p_\lambda$ (only for ABCEI)& $\{10^k\}_{k=-5}^3$\\
						$p_\beta$ (only for ABCEI)& $\{10^k\}_{k=-5}^3$\\
						Num. shared layers &  $\{1,2,3\}$\\
						Num. representation layers &  $\{1,2,3\}$\\
						Dim. shared layers (NEWS) & $\{500,250\}$\\
						Dim. representation layers (NEWS)&  $\{100,50\}$\\
						Dim. shared layers (ACIC2016) & $\{20,50,100,200\}$\\
						Dim. representation layers (ACIC2016)&  $\{20,50,100,200\}$\\
						Dropout &  $\{0.0,0.3,0.5\}$\\
						Batch Size (NEWS) &$\{315,525\}$\\
						Batch Size (ACIC2016) &$\{100,200\}$\\
						\bottomrule
					\end{tabular}
	\end{table}
	
	\paragraph{SITE} We used the original SITE implementation of \citet{Yao2018} obtained from \url{https://github.com/Osier-Yi/SITE}. For IHDP, SITE uses the same architecture as the best proposed CFRNet and TARNet (described above) and we evaluated $10$ randomly sampled parameter combinations for the strength of the Position Dependent Deep Metric (PDDM) and Middle Point Distance Minimization (MPDM), as listed in \cref{Stab:CFRparameters}. For NEWS and ACIC2016, we randomly sampled $25$ parameter sets given the set of hyper-parameters outlined in \cref{Stab:CFRparameters}. We selected the best model based on minimal factual MSE following \citet{Yao2018}.
	
	\paragraph{ABCEI} We used the original implementation of ABCEI by \citet{du2021adversarial} obtained from \textit{https://github.com/octeufer/Adversarial-Balancing-based-representation-learning-for-Causal-Effect-Inference}. For IHDP, we used the proposed set of hyper-parameters, i.e., $3$ input layers of size $200$, 3 output layers of size $100$, ELU activation functions, a batch size of $65$ $p_\lambda=0.0001$ and $p_\beta=10$. The hyper-parameters creened for NEWS and ACIC2016 are listed in \cref{Stab:CFRparameters}.
	
	\begin{table}
		\caption{Hyper-parameter ranges for (U)ADBCR and DANNCR}
		\label{Stab:ADBCR_parameters}
		\centering
					\begin{tabular}{ll}
					    \toprule
					    Parameter & Range\\
						\midrule
						shared layers (IHDP, ACIC2016)& $\{[50,50], [20, 20], [10,10]\}$\\
						ind. layers (IHDP, ACIC2016)& $\{[50,50], [20,20], [10]\}$\\
						shared layers (NEWS)& $\{[500, 250], [250, 100]\}$\\
						ind. layers (NEWS)& $\{[100,50], [50,10]\}$\\
						Dropout &  $\{0.1,0.3,0.5\}$\\
						Weight Decay & $\{1, 0.1, 0.01, 0.001\}$\\
						Batch Size (IHDP, ACIC2016) &$\{100, 250, 500\}$\\
						Batch Size (NEWS) &$\{525, 1575, 3150\}$\\
						learning rate & $[10^{-5},10^-2]$ \\
						Num. adv. Steps $k$ &$\{1,2,3\}$\\
						\bottomrule
					\end{tabular}
	\end{table}

	\paragraph{(U)ADBCR} (U)ADBCR is publicly available at \url{GitHub(censored)} (\textit{See additional Supplementary Material}).
	The Hyper-parameter ranges used for (U)ADBCR are listed in \cref{Stab:ADBCR_parameters} with different network architectures and batch sizes for the IHDP and NEWS dataset, respectively. In contrast to CFRNet and SITE, ADBCR allows for custom layer sizes for both the latent representation and the outcome heads. Hence, the dimensions for consecutive layers are given in squared brackets. We further add a dropout layer after each dense layer, used ELU activation functions, and optimize the parameters using Adam \cite{Kingma2014}.
	
	\paragraph{DANNCR} Based on the ideas by \cite{Ozery-Flato2018} and \cite{bica2020estimating}, we implemented DANNCR which uses a DANN based adversary \cite{ganin2016domain} to align treatment and control distributions. DANNCR uses a similar architecture like (U)ADBCR but with a single outcome head per treatment. The additional domain discriminator uses the same architecture as the respective outcome heads with two outcome values and optimizes a cross-entropy objective to classify whether the samples were generated by the treated or untreated population. For model selection, we use the same hyper-parameter space as for (U)ADBCR (\cref{Stab:ADBCR_parameters}) and select the best DANNCR model based on factual accuracy.
	
	\begin{table}[t]
		\caption{Comparison between (U)ADBCR and the results reported by \citet{Shalit2017} evaluated on $1000$ realizations of the IHDP data.}
        \centering
        \label{Stab:Shalit}
					\begin{tabular}{lrrrr}
						\toprule
						&\multicolumn{2}{c}{Within-Sample} 						  & \multicolumn{2}{c}{Out-of-sample}\\
						&$\sqrt{\epsilon_{\text{PEHE}}}$ & $\epsilon_{\text{ATE}}$ & $\sqrt{\epsilon_{\text{PEHE}}}$ & $\epsilon_{\text{ATE}}$ \\
						\midrule
						OLS-S& 5.8 $\pm$ .3 & .73 $\pm$ .04 & 5.8 $\pm$ .3 & .94 $\pm$ .06\\
						OLS-T& 2.4 $\pm$ .1 & .14 $\pm$ .01 & 2.5 $\pm$ .1 & .31 $\pm$ .02\\
						BLR& 5.8 $\pm$ .3 & .72 $\pm$ .04 & 5.8 $\pm$ .3 & .93 $\pm$ .05\\
						k-NN& 2.1 $\pm$ .1 & .14 $\pm$ .01 & 4.1 $\pm$ .2 & .79 $\pm$ .05\\
						TMLE& 5.0 $\pm$ .2 & .30 $\pm$ .01 & --- & ---\\
						BART& 2.1 $\pm$ .1 & .23 $\pm$ .01 & 2.3 $\pm$ .1 & .34 $\pm$ .02\\
						R.For.& 4.2 $\pm$ .2 & .73 $\pm$ .05 & 6.6 $\pm$ .3 & .96 $\pm$ .06\\
						C.For.& 3.8 $\pm$ .2 & .18 $\pm$ .01 & 3.8 $\pm$ .2 & .40 $\pm$ .03\\
						BNN& 2.2 $\pm$ .1 & .37 $\pm$ .03 & 2.1 $\pm$ .1 & .42 $\pm$ .03\\
						\midrule
						TARNet & .88 $\pm$ .02 & .26 $\pm$ .01 & .95 $\pm$ .02 & .28 $\pm$ .01\\
						CFR MMD& .73 $\pm$ .01 & .30 $\pm$ .01 & .78 $\pm$ .02 & .31 $\pm$ .01\\
						CFR Wass.& .71 $\pm$ .02 & .25 $\pm$ .01 & .76 $\pm$ .02 & .27 $\pm$ .01\\
						\midrule
						\bfseries ADBCR & $.71\pm .02$ & $.16\pm .01$ & $.79\pm .03$ & $.18\pm .01$\\
						\bfseries UADBCR & $.70\pm .02$ & $.16\pm .01$ & $.78\pm .03$ & $.17\pm .01$\\
						\bottomrule
					\end{tabular}
	\end{table}	
	
	We performed a grid-search over the different architectures and randomly sampled the other hyper parameters for $30$ network initializations. Note that the network initialization is important to find an appropriate set of adjacent representations. To reduce computational burden, we stopped optimization for a non-increasing validation loss of $100$ consecutive epochs. Finally, we used the stopping criterion to identify the best model for each data representation.

	\section*{Additional IHDP performance benchmarks}
	The IHDP benchmark has been used in a variety of different settings, comprising different realizations of the simulation parameters and a scaling of the outcome values. These differences, however, often prevent a fair comparison between state-of-the-art models. Therefore, we have additionally evaluated (U)ADBCR for the first $10$ (SITE), $100$ realizations with normalized outcome values (Meta-Learners), and the full $1000$ realizations of the IHDP data. The results comparing (U)ADBCR to the original reported results are shown in \cref{Stab:Shalit}, \cref{Stab:SITE} and \cref{Stab:IHDP_meta}, respectively, including additional non-deep learning baselines. For the full details, we refer to the respective original papers \cite{Shalit2017,Curth2021a,Yao2018}.

	\begin{table}
		\caption{Comparison between (U)ADBCR and the results reported by \citet{Yao2018} for the first $10$ realizations of the IHDP data.}
		\label{Stab:SITE}
		\centering
					\begin{tabular}{lrr}
						\toprule
						&Within-Sample 						  & Out-of-sample\\
						&$\sqrt{\epsilon_{\text{PEHE}}}$ &  $\sqrt{\epsilon_{\text{PEHE}}}$ \\
						\midrule
						OLS/LR1 & $10.761 \pm 4.350$ & $7.345 \pm 2.914$ \\
						OLS/LR2 &$10.280 \pm 3.794$ & $5.245 \pm 0.986$  \\
						HSIC-NNM &$2.439 \pm 0.445$ & $2.401 \pm 0.367$ \\
						PSM  &$7.188 \pm 2.679$ &$ 7.290 \pm 3.389$ \\
						k-NN & $4.432 \pm 2.345$ & $4.303 \pm 2.077$ \\
						Causal Forest & $4.732 \pm 2.974$ & $4.095 \pm 2.528$\\ 
						BNN & $3.827 \pm 2.044$ & $4.874 \pm 2.850$ \\
						TARNet & $0.729 \pm 0.088$ & $1.342 \pm 0.597$\\ 
						CFR-MMD & $0.663 \pm 0.068$ &$ 1.202 \pm 0.550$ \\
						CFR-WASS & $0.649 \pm 0.089$ & $1.152 \pm 0.527$ \\
						SITE & $0.604 \pm 0.093$ & $0.656 \pm 0.108$ \\
						\midrule
						\bfseries ADBCR & $0.603 \pm 0.208$ &  $0.595 \pm 0.175$ \\
						\bfseries UDA-ADBCR & $0.562 \pm 0.160$ & $0.543 \pm 0.121$\\
						\bottomrule
					\end{tabular}
	\end{table}

	\begin{table}[t]
		\caption{Comparison between (U)ADBCR and the results reported by \citet{Curth2021a} for $100$ realizations of the adopted IHDP data.}
		\label{Stab:IHDP_meta}
		\vskip 0.15in
		\begin{small}
			\begin{center}
				\begin{sc}
					\begin{tabular}{l|r|r}
						\toprule
						&Within-Sample						  &Out-of-sample\\
						&$\sqrt{\epsilon_{\text{PEHE}}}$  & $\sqrt{\epsilon_{\text{PEHE}}}$ \\
						\midrule
						TNet &0.761 (0.011)& 0.770 (0.013)\\
						SNet-1& 0.678 (0.009)& 0.689 (0.012)\\
						SNet-2& 0.676 (0.009)& 0.687 (0.012)\\
						SNet-3& 0.683 (0.009)& 0.691 (0.011)\\
						SNet& 0.729 (0.009)& 0.737 (0.011)\\
						RA-L. + TNet& 0.740 (0.010)& 0.745 (0.013)\\
						RA-L. + SNet-2& 0.670 (0.009)& 0.680 (0.012)\\
						DR-L. + TNet& 0.893 (0.017)& 0.902 (0.019)\\
						PW-L. + TNet& 2.250 (0.093)& 2.285 (0.094)\\
						\midrule
						\bfseries ADBCR & 0.412 (0.012) & 0.391 (0.009) \\
						\bfseries UADBCR & 0.425 (0.014) & 0.407 (0.013) \\
						\bottomrule
					\end{tabular}
				\end{sc}
			\end{center}
		\end{small}
		\vskip -0.1in
	\end{table}

\end{document}